%% file: main.tex
\input{constants}
\arxiv %

\pdfoutput=1
\documentclass[10pt,twocolumn,letterpaper]{article}
\input{cvpr_header}
\begin{document}
\title{\paperTitle}
\author{\authorBlock}

\twocolumn[{%
\renewcommand\twocolumn[1][]{#1}%
\maketitle
\input{figures/introduction/teaser}
}]

{
  \renewcommand{\thefootnote}%
    {\fnsymbol{footnote}}
  \footnotetext[1]{Equal contribution}
}

\input{00_abstract}
\input{01_intro}

\input{02_related}
\input{03_method}

\input{04_implementation}
\input{05_experiments}

\input{06_limitations}

\input{10_conclusion}

{\small
    \bibliographystyle{ieee_fullname}
    \bibliography{11_references}
}

\end{document}

%% file: constants.tex
\def\cvprPaperID{9506}
\def\confName{CVPR}
\def\confYear{2023}

\def\paperTitle{Learning Neural Volumetric Representations of Dynamic Humans in Minutes}

\def\authorBlock{
    Chen Geng\footnotemark[1] \qquad
    Sida Peng\footnotemark[1] \qquad
    Zhen Xu\footnotemark[1] \qquad
    Hujun Bao \qquad
    Xiaowei Zhou \\
    State Key Laboratory of CAD\&CG, Zhejiang University \\
}

\newif\ifreview 
\newif\ifarxiv \newcommand{\arxiv}{\arxivtrue}
\newif\ifcamera 
\newif\ifrebuttal 

%% file: cvpr_header.tex
\ifreview \usepackage[review]{cvpr} \fi
\ifarxiv \usepackage[pagenumbers]{cvpr} \fi
\ifrebuttal \usepackage[rebuttal]{cvpr} \fi
\ifcamera \usepackage{cvpr} \fi

\usepackage{graphicx}
\usepackage{amsmath}
\usepackage{amssymb}
\usepackage{booktabs}

\input{macros}  %

\usepackage{xr-hyper}

\makeatletter
\newcommand*{\addFileDependency}[1]{
  \typeout{(#1)}
  \@addtofilelist{#1}
  \IfFileExists{#1}{}{\typeout{No file #1.}}
}

\makeatother

\usepackage[pagebackref,breaklinks,colorlinks]{hyperref}
\usepackage[capitalize]{cleveref}
\crefname{section}{Sec.}{Secs.}
\crefname{table}{Table}{Tables}
\crefname{figure}{Fig.}{Figs.}
\newcommand{\rb}[1]{{#1}}
\DeclareMathOperator*{\argmax}{argmax}

\usepackage[tableposition=top]{caption}
\usepackage{multirow}
\usepackage[table,xcdraw]{xcolor}
\usepackage{lscape}
\usepackage{wrapfig,lipsum,booktabs}

\newcolumntype{x}[1]{>{\centering\arraybackslash}p{#1pt}}

\newlength\savewidth\newcommand\shline{\noalign{\global\savewidth\arrayrulewidth
  \global\arrayrulewidth 1pt}\hline\noalign{\global\arrayrulewidth\savewidth}}

\newcommand{\revise}[1]{#1}

\newcommand\rowtopspace{\rule{0pt}{1.3em}}       %
\newcommand\rowbotspace{\rule[-0.75em]{0pt}{0pt}} %

%% file: macros.tex
\usepackage{times}
\usepackage{microtype}
\usepackage{epsfig}
\usepackage[table,xcdraw]{xcolor}
\usepackage{caption}
\usepackage{float}
\usepackage{placeins}
\usepackage{color, colortbl}
\usepackage{stfloats}
\usepackage{enumitem}
\usepackage{tabularx}
\usepackage{xstring}
\usepackage{multirow}
\usepackage{xspace}
\usepackage{url}
\usepackage{subcaption}
\usepackage{xcolor}
\usepackage[hang,flushmargin]{footmisc}

\ifcamera \usepackage[accsupp]{axessibility} \fi

\ifarxiv  \fi

\newcommand{\R}[1]{{%
    \textbf{%
        \ifstrequal{#1}{1}{\textcolor{red}{R#1}}{%
        \ifstrequal{#1}{2}{\textcolor{blue}{R#1}}{%
        \ifstrequal{#1}{3}{\textcolor{magenta}{R#1}}{%
        \ifstrequal{#1}{4}{\textcolor{teal}{R#1}}{%
                           \textcolor{cyan}{R#1}%
        }}}}%
    }%
}}

%% file: figures/introduction/teaser.tex
\begin{center}
    \centering
    \includegraphics[width=1\linewidth]{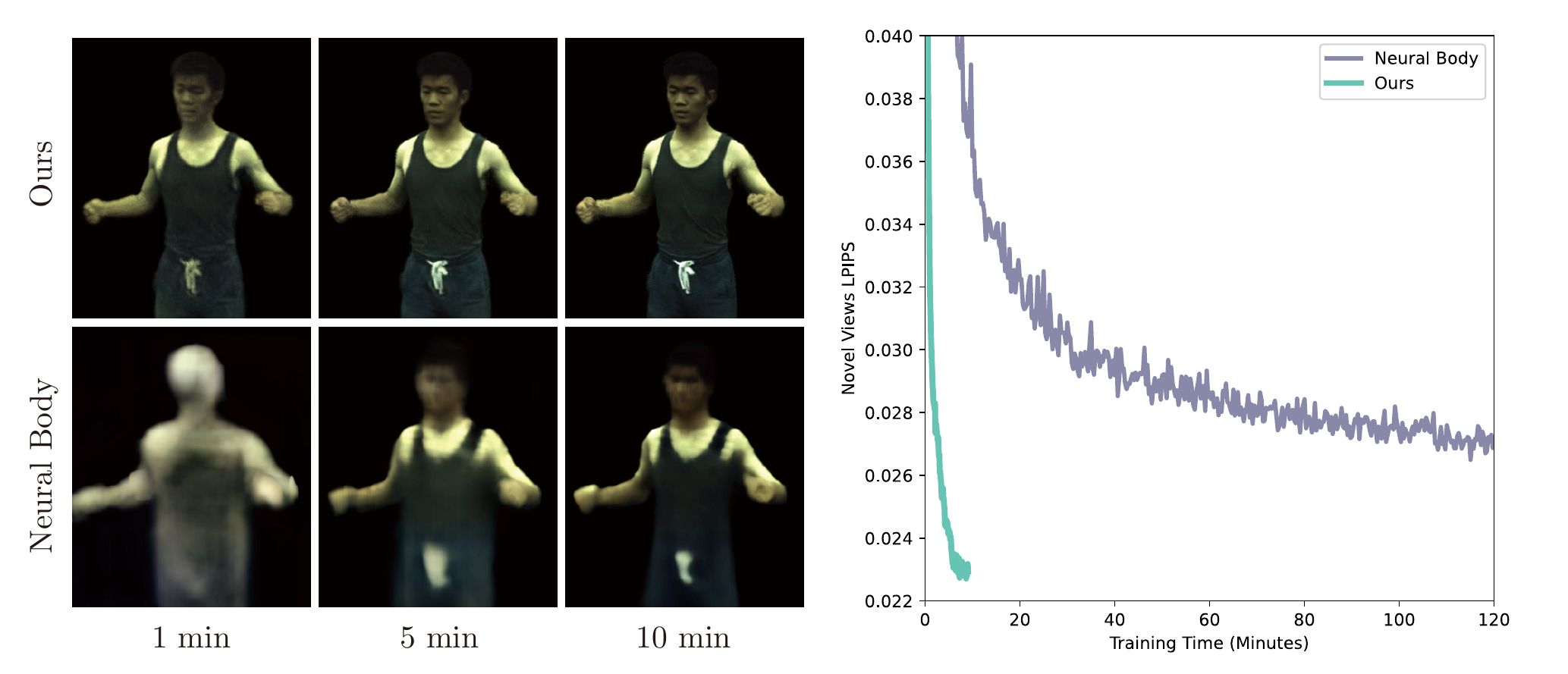}
    \captionsetup{type=figure}
    \vspace{-1.5em}
    \captionof{figure}{\textbf{Convergence rate of training.} Given a monocular video of a human performer, our model can be learned in $\sim$5 minutes to produce photorealistic novel view rendering, which is 100 times faster than Neural Body \cite{peng2021neural}.}
    \vspace{1.5em}
    \label{fig:teaser}
\end{center}

%% file: 00_abstract.tex
\begin{abstract}
    This paper addresses the challenge of quickly reconstructing free-viewpoint videos of dynamic humans from sparse multi-view videos.
    Some recent works represent the dynamic human as a canonical neural radiance field (NeRF) and a motion field, which are learned from videos through differentiable rendering. But the per-scene optimization generally requires hours.
    Other generalizable NeRF models leverage learned prior from datasets and reduce the optimization time by only finetuning on new scenes at the cost of visual fidelity.
    In this paper, we propose a novel method for learning neural volumetric videos of dynamic humans from sparse view videos in minutes with competitive visual quality.
    Specifically, we define a novel part-based voxelized human representation to better distribute the representational power of the network to different human parts.
    Furthermore, we propose a novel 2D motion parameterization scheme to increase the convergence rate of deformation field learning.
    Experiments demonstrate that our model can be learned 100 times faster than prior per-scene optimization methods while being competitive in the rendering quality. Training our model on a $512 \times 512$ video with 100 frames typically takes about 5 minutes on a single RTX 3090 GPU. The code will be released on our project page: \href{https://zju3dv.github.io/instant_nvr}{https://zju3dv.github.io/instant\_nvr}.
\end{abstract}

%% file: 01_intro.tex
\section{Introduction}
\label{sec:intro}

Creating volumetric videos of human performers has many applications, such as immersive telepresence, video games, and movie production.
Recently, some methods \cite{peng2021neural, weng2022humannerf} have shown that high-quality volumetric videos can be recovered from sparse multi-view videos by representing dynamic humans with neural scene representations.
However, they typically require more than 10 hours of training on a single GPU.
The expensive time and computational costs limit the large-scale application of volumetric videos.
Generalizable methods \cite{yu2021pixelnerf, kwon2021neural} utilize learned prior from datasets of dynamic humans to reduce the training time by only finetuning on novel human performers. These techniques could increase the optimization speed by a factor of 2-5 at the cost of some visual fidelity.

To speed up the process of optimizing a neural representation for view synthesis of dynamic humans, we analyze the structural prior of the human body and motion and propose a novel dynamic human representation that achieves 100x speedup during optimization while maintaining competitive visual fidelity.
Specifically, to model a dynamic human, we first transform world-space points to a canonical space using a novel motion parameterization scheme and inverse linear blend skinning (LBS) \cite{lewis2000pose}.
Then, the color and density of these points are estimated using the canonical human model.

The innovation of our proposed representation is two-fold. First, we observe that different human parts have different shape and texture complexity. For example, the face of a human performer typically exhibits more complexity than a flatly textured torso region, thus requiring more representational power to depict. Motivated by this, our method decomposes the canonical human body into multiple parts and represents the human body with a structured set of voxelized NeRF \cite{mildenhall2020nerf} networks to bring the convergence rate of these different parts to the same level.
In contrast to a single-resolution representation, the part-based body model utilizes the human body prior to represent the human shape and texture efficiently, heuristically distributing variational representational power to human parts with spatially varying complexity.

Second, we notice that human motion typically occurs around a surface instead of in a volume, that is, near-surface points that are projected to neighboring regions on a parametric human model have similar motion behavior. Thus we propose a novel motion parameterization technique that models the 3D human deformation in a 2D domain. This idea is similar to the displacement map and bump map \cite{cook1987reyes, cook1984shade} in traditional Computer Graphics to represent detailed deformation on a 2D texture domain. We extend the technique of displacement map \cite{cook1987reyes, cook1984shade} to represent human motions by restricting the originally 3D deformation field \cite{pumarola2021d, liu2021neural, peng2021animatable} onto the 2D surface of a parametric human model, such as SMPL \cite{loper2015smpl}.
This technique significantly increases the convergence rate of a deformation field by reducing the dimensionality at which the neural representation needs to model the motion.

Experiments demonstrate that our method significantly accelerates the optimization of neural human representations while being competitive with recent human modeling methods on the rendering quality.
As shown in Figure~\ref{fig:teaser}, our model can be trained in around 5 minutes to produce a volumetric video of a dynamic human from a 100-frame monocular video of $512 \times 512$ resolution on an RTX 3090 GPU.

To summarize, our key contributions are:
\begin{itemize}
    \setlength{\itemsep}{1pt}
    \setlength{\parskip}{0pt}
    \setlength{\parsep}{0pt}
    \item A novel part-based voxelized human representation for more efficient human body modeling.
    \item A 2D motion parameterization scheme to for more efficient deformation field modeling.
    \item 100x speedup in optimization compared to previous neural human representations while maintaining competitive rendering quality.
\end{itemize}

%% file: 02_related.tex
\input{figures/method/pipeline}

\section{Related work}
\label{sec:related}

\paragraph{Implicit neural representation and rendering.} There have been many 3D scene representations, such as multi-view images \cite{su2015multi, qi2016volumetric}, textured meshes \cite{thies2019deferred, liao2020towards}, point clouds \cite{raj2021anr, aliev2020neural}, and voxels \cite{sitzmann2019deepvoxels, lombardi2019neural}. Recently, some methods \cite{sitzmann2019scene, chen2019learning, ye2022gifs, mescheder2019occupancy, park2019deepsdf, liu2020dist, zhi2021place} propose implicit neural representations to represent scenes, which uses MLP networks to predict scene properties for any point in 3D space, such as occupancy \cite{mescheder2019occupancy, saito2019pifu}, signed distance \cite{park2019deepsdf, liu2020dist}, and semantics \cite{zhi2021place, guo2022manhattan}. This enables them to describe continuous and high-resolution 3D scenes.
To perform novel view synthesis, neural radiance field (NeRF) \cite{mildenhall2020nerf} models scenes as implicit fields of density and color. NeRF is optimized from images with volume rendering techniques, which produces impressive image synthesis results. Many works improve NeRF in various aspects, including rendering quality \cite{barron2021mip, barron2021mip360}, rendering speed \cite{reiser2021kilonerf, liu2020neural, yu2021plenoctrees, hedman2021baking, garbin2021fastnerf}, scene scale \cite{tancik2022block, turki2021mega, xiangli2021citynerf, rematas2021urban}, and reconstruction quality \cite{yariv2021volume, wang2021neus, oechsle2021unisurf}. Some methods \cite{park2021nerfies, pumarola2021d, du2021neural, li2021neural} extend NeRF to dynamic scenes.

\paragraph{Human modeling.} Reconstructing high-quality 3D human models is essential for synthesizing free-viewpoint videos of human performers. Traditional methods leverage multi-view stereo techniques \cite{schonberger2016structure, schonberger2016pixelwise, guo2019relightables} or depth fusion \cite{su2020robustfusion, collet2015high, dou2016fusion4d} to reconstruct human geometries, which require complicated hardware, such as dense camera arrays or depth sensors. To reduce the requirement of the capture equipment, some methods \cite{saito2019pifu, saito2020pifuhd, alldieck2022photorealistic} train networks to learn human priors from datasets containing a large amount of 3D ground-truth human models, enabling them to infer human geometry and texture from even a single image. However, due to the limited diversity of training data, these methods do not generalize well to humans under complex poses.
Recently, some methods \cite{saito2021scanimate, chen2021snarf, tiwari2021neural, mihajlovic2022coap} model the shapes of dynamic humans as implicit neural representations and attempt to optimize them from human scans. 
Another line of works \cite{peng2021neural, kwon2021neural, xu2021h, zheng2022structured, zhao2021humannerf, weng2022humannerf, jiang2022selfrecon, peng2021animatable, liu2021neural, zhang2022ndf, li2022tava, wang2022arah, zhi2022dual, remelli2022drivable, hu2021hvtr} exploits dynamic implicit neural representations and differentiable renderers to reconstruct 3D human models from videos. To represent dynamic humans, Neural Actor \cite{liu2021neural} augments the neural radiance field with the linear blend skinning model \cite{lewis2000pose}. It additionally adopts a residual deformation field to better predict human motions.
To overcome the inaccuracy of input human poses, \cite{su2021nerf, weng2022humannerf} optimize the parameters of human poses jointly with the human representations during training.
These methods typically require a lengthy training process to produce high-quality human models.
In contrast, we introduce a part-based voxelized human representation to model the canonical human body, which significantly accelerates the optimization process.
Although \cite{deng2020nasa, mihajlovic2022coap} have proposed part-based implicit functions, they focus on human shape modeling and do not show that the part-based representation can be used to reduce the training time.

\paragraph{Accelerating the optimization of neural representations.} Many differentiable rendering-based methods \cite{mildenhall2020nerf, liu2020neural, barron2021mip} optimize a separate neural representation for each scene. The optimization process generally takes several hours on a modern GPU, which is time-consuming and costly to scale. Inspired by multi-view stereo matching \cite{schonberger2016pixelwise}, some methods \cite{wang2021ibrnet, chen2021mvsnerf, chibane2021stereo, trevithick2021grf, shao2021doublefield, yu2021pixelnerf, mihajlovic2022keypointnerf} train a network on multi-view datasets to learn to infer radiance fields from input images. This enables them to quickly fine-tune neural representations to unseen scenes. \cite{ramon2021h3d, dupont2022data} leverage the auto-decoder \cite{park2019deepsdf} to capture the scene priors for efficient fine-tuning. \cite{bergman2021fast, tancik2021learned} utilize meta-learning techniques \cite{nichol2018first, finn2017model} to initialize network parameters, thereby improving the training speed.
Some methods \cite{saragadam2022miner, yu2021plenoxels, chen2022tensorf, sun2021direct, peng2022selfnerf, chen2022uv} attempt to design scene representations that support efficient training.
\cite{tancik2020fourier, sitzmann2020implicit, muller2022instant, gao2022reconstructing} augments the approximation ability of networks by designing encoding techniques. Multiresolution hash encoding \cite{muller2022instant} defines multiresolution feature vector arrays for a scene and uses the hash technique \cite{teschner2003optimized} to assign each input coordinate a feature vector as the encoded input, which significantly improves the training speed.

%% file: figures/method/pipeline.tex
\begin{figure*}[t]
    \centering
    \includegraphics[width=1.0\linewidth]{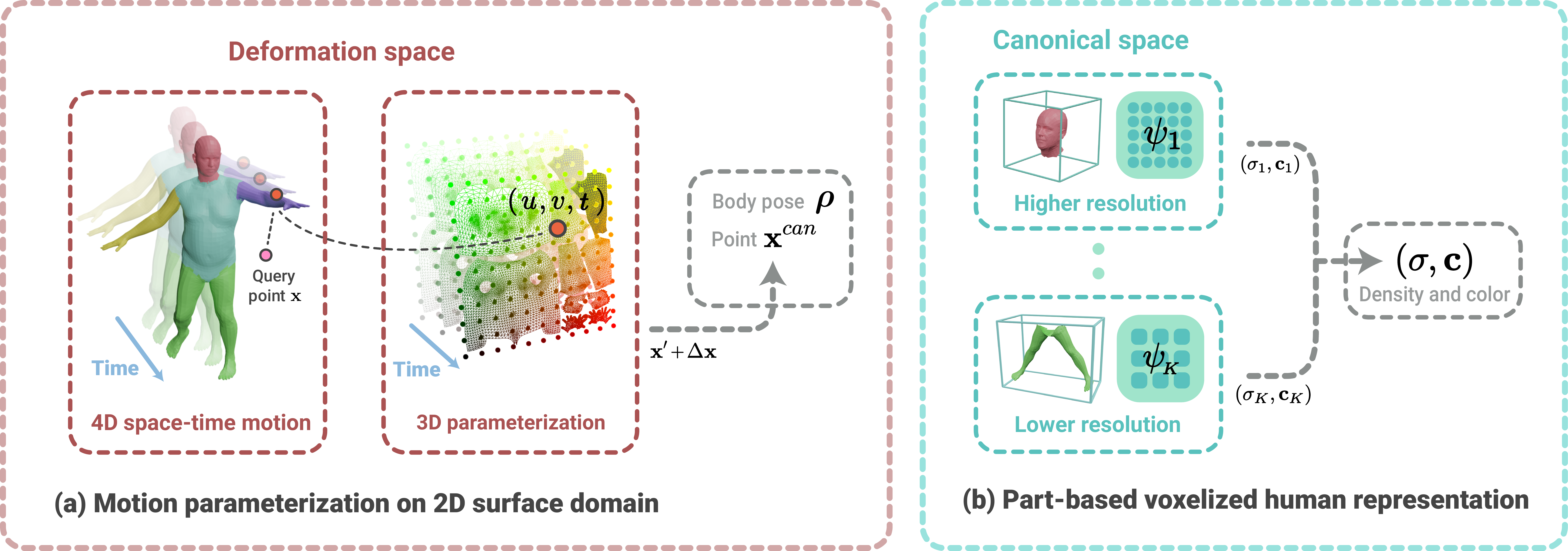}
    \vspace{-0.5em}
    \caption{\textbf{Overview of the proposed representation.} Given a query point $\mathbf{x}$ at frame $t$, we find its nearest surface point on each human part of the SMPL mesh, which gives the blend weight $\mathbf{w}_k$ and the UV coordinate $(u_k, v_k)$.
    Consider the $k$-th part. The motion field consists of an inverse LBS module and a residual deformation module. (a) The inverse LBS module takes body pose $\boldsymbol{\rho}$, blend weight $\mathbf{w}_k$, and query point $\mathbf{x}$ as input and output the transformed point $\mathbf{x}'$. The residual deformation module applies the multiresolution hash encoding (MHE) to $(u_k, v_k, t)$ and uses an MLP network to regress the residual translation $\Delta \mathbf{x}$, which is added to $\mathbf{x}'$ to obtain the canonical point $\mathbf{x}^{\text{can}}$.
    (b) We then feed $\mathbf{x}^{\text{can}}$ to networks of $k$-th human part to predict the density $\sigma_k$ and color $\mathbf{c}_k$.
    With $\{(\sigma_k, \mathbf{c}_k)\}_{k=1}^K$, we select the one with the biggest density as the density and color of the query point.
    }
    \label{fig:pipeline}
\end{figure*}

%% file: 03_method.tex
\section{Method}
\label{sec:method}

This paper aims to quickly create a 3D video from a sparse multi-view video that captures a dynamic human. Following Neural Actor \cite{liu2021neural}, we assume that the cameras are calibrated, and the human pose and foreground human mask of each image are provided.

In this section, we build a dynamic human model that is comprised of a part-based voxelized human representation and a dimensionality reduction motion parameterization scheme (Section~\ref{par:part-based}).
Then, Section~\ref{sec:training} discusses how to efficiently optimize the proposed representation.
Finally, we provide implementation details in Section~\ref{sec:implementation}.

\subsection{Proposed human representation}
\label{sec:representation}

As shown in Figure~\ref{fig:pipeline}, our dynamic human representation consists of a motion parameterization field and a part-based voxelized human model.
(a) For a query point $\mathbf{x}$, the motion parameterization field first transforms it to the canonical space correspondence $\mathbf{x}^{can}$ using the inverse LBS \cite{lewis2000pose} algorithm and by parameterizing the 3D points onto 2D UV coordinates to predict the residual deformation $\Delta\mathbf{x}$.
(b) Then, $\mathbf{x}^{can}$ is fed into the part-based voxelized human model in canonical space to predict and aggregate the density and color $(\sigma, \mathbf{c})$, where the canonical human body is decomposed into $K$ parts, each of which is represented using an MHE-augmented \cite{muller2022instant} NeRF network.

\paragraph{Motion parameterization on 2D surface domain.}
\label{par:motion-parameterization}

To regress the canonical correspondence $\mathbf{x}^{can}$ of a query point $\mathbf{x}$, we first find its nearest surface point $\mathbf{p}$ on the posed SMPL mesh.
Using the strategy in \cite{liu2021neural}, the blend weight $\mathbf{w}$ and UV coordinate $(u, v)$ of surface point $\mathbf{p}$ are obtained from the SMPL model.

Given the blend weight $\mathbf{w}$ and UV coordinate $(u, v)$, the motion field maps the query point $\mathbf{x}$ to the canonical space correspondence $\mathbf{x}^{can}$. The motion field is comprised of an inverse LBS module \cite{lewis2000pose} and a residual deformation module. Given a query point $\mathbf{x}$ and blend weight $\mathbf{w}$, we use the inverse LBS module to transform it to the unposed space, which is defined as:
\begin{equation}
    \Phi_{\text{LBS}}(\mathbf{x}, \mathbf{w}, \boldsymbol{\rho}) = \left( \sum_{j=1}^J w_{k,j} G_j \right)^{-1} \mathbf{x},
\end{equation}
where $\boldsymbol{\rho}$ denotes the human pose and $\{G_j\}_{j=1}^J$ are transformation matrices derived from $\boldsymbol{\rho}$ \cite{loper2015smpl}. The detailed derivation of the inverse LBS algorithm can be found in the supplementary material.

The transformed point $\Phi_{\text{LBS}}(\mathbf{x}, \mathbf{w}, \boldsymbol{\rho})$ is then deformed to the human surface using the residual deformation module.
Specifically, the current time $t$ is first concatenated with the UV coordinate $(u, v)$ to serve as the parameterization of the query point $\mathbf{x}$ at frame $t$. This motion parameterization is inspired by the displacement map and bump map techniques \cite{cook1987reyes, cook1984shade} in the traditional Computer Graphics pipelines. It essentially reduces the dimensionality of a 4D space-time sequence down to the 3D surface-time domain utilizing the human deformation prior. Then, we apply the multiresolution hash encoding \cite{muller2022instant} $\psi_{\text{res}}$ to $(u, v, t)$ and forward the encoded input through a network $\text{MLP}_{\text{res}}$ to regress the residual $\delta$. The full human motion at frame $t$ is defined as:
\begin{align}
    \Delta\Phi(u, v, t) = \text{MLP}_{\text{res}}(\psi_{\text{res}}(u, v, t)), \\
    \begin{aligned}
        \Phi(\mathbf{x}, \mathbf{w}, u, v, \boldsymbol{\rho}, t) = \Phi_{\text{LBS}}(\mathbf{x}, \mathbf{w}, \boldsymbol{\rho}) \\
        + \Delta\Phi(u, v, t).
    \end{aligned}
\end{align}

There are two main observations that inspired us to use the $(u, v, t)$ motion parameterization.
First, we observe that a typical human motion happens at a surface level instead of a volumetric level. Near-surface points sharing similar UV coordinate of the parametric model shows similar motions. Utilizing this prior with a surface parameterization \cite{cook1987reyes, cook1984shade}, we can reduce the required 4D volumetric motion to the 3D surface-time domain, greatly decreasing the amount of information the deformation network has to learn.
Based on a similar idea, \cite{bhatnagar2020loopreg} diffuses the surface motion to the full 3D space.
Second, a naive $(x, y, z, t)$ encoding would introduce quartic memory overhead on an explicitly defined voxel structure, which is intractable to use in practice. Instead, by parameterizing the motion to $(u, v, t)$, we can reduce the memory footprint to a more practical cubic level.
Experiments demonstrate that the motion parameterization scheme effectively reduces the dimensionality of the deformation field, thus greatly increasing the convergence rate of the human model.

\paragraph{Part-based voxelized human representation.}
\label{par:part-based}

Muller et al. \cite{muller2022instant} propose the multi-resolution hash encoding (MHE) to improve the approximation ability and training speed of implicit neural representations. MHE is defined on an explicit set of voxel grids of different resolutions. Given an input coordinate, it applies the hash encoding on each level and queries the corresponding voxel grid to trilinearly interpolate the feature of the input point for this level. Then, the concatenated multi-resolution feature is fed into a small MLP network to predict the target value. Note that \cite{muller2022instant} concatenates the features of multi-resolution hash encoding for the same point to mitigate the effect of hash collision, while our part-based voxelized human representation introduces spatially varying resolution to efficiently encode human parts with different complexity.

In contrast to \cite{peng2021animatable, liu2021neural} which use a single neural radiance field (NeRF) to represent the canonical human model, we decompose the human body into multiple parts with different complexity and adopt a structured set of MHE-augmented NeRF with varying resolutions as the body representation.
Specifically, we manually divide the human body into multiple parts based on a parametric human model (such as SMPL \cite{loper2015smpl}), as shown in Figure~\ref{fig:pipeline}. Note that other parametric human models \cite{romero2017embodied, pavlakos2019expressive} can also be used in our method. We use the blend weights defined in SMPL model \cite{loper2015smpl} to decompose SMPL template mesh $\mathcal{M} = (\mathcal{V}, \mathcal{E})$, where $\mathcal{V}$ represents the vertices and $\mathcal{E}$ represents the edges. Let the $i$-th vertice $v_i$ have blend weight $w_i$ and for each part $k$ we define $\Omega_k$ as the set of bones that belong to this part. The detailed setting of $\Omega_k$ can be found in the supplementary material. The mesh of the $k$-th part is defined as $\mathcal{M}_k = (\mathcal{V}_k, \mathcal{E}_k)$, where:

\begin{align}
    \mathcal{V}_k = \{v_i | \argmax{w_i} \in \Omega_k\},
    \\
    \mathcal{E}_k = \{(v_i, v_j) | v_i \in \mathcal{V}_k, v_j \in \mathcal{V}_k\}.
\end{align}

To regress the density and color of a query point $\mathbf{x}$, we first find the nearest surface point $\mathbf{p}_k$ on each human part \rb{$\mathcal{M}_k$} of the posed SMPL mesh.
Using the strategy in \cite{liu2021neural}, the blend weight $\mathbf{w}_k$ and UV coordinate $(u_k, v_k)$ of surface point $\mathbf{p}_k$ are obtained from the SMPL model.
With $(u_k, v_k, t)$, we use the motion parameterization scheme defined in Section~\ref{par:motion-parameterization} to transform the query point to the space of the $k$-th human part. We predefine the parameters of the multiresolution hash encoding function $\psi_k$ for the $k$-th part. Given the transformed point, we first apply the multiresolution hash encoding to the transformed point and then feed the encoded point $\psi_k(\mathbf{x})$ to a small NeRF network to predict the density and color. The density network $\text{MLP}_{\sigma_k}$ is defined as:
\begin{equation}
    (\sigma_k, \mathbf{z}) = \text{MLP}_{\sigma_k}(\psi_k(\mathbf{x})),
\end{equation}
where $\sigma_k$ means the density and $\mathbf{z}$ is a feature vector. Then, we take the feature vector $\mathbf{z}$ and the viewing direction $\mathbf{d}$ as the input for the color regression. Similar to \cite{park2021nerfies}, a latent embedding $\boldsymbol{\ell}_t$ for each video frame $t$ is introduced to model the temporally-varying appearance. The color network is defined as:
\begin{equation}
    \mathbf{c}_k = \text{MLP}_{\mathbf{c}_k}(\mathbf{z}, \mathbf{d}, \boldsymbol{\ell}_t).
\end{equation}
Finally, we have $K$ predictions $\{(\sigma_k, \mathbf{c}_k)\}_{k=1}^K$. The density and color $(\sigma, \mathbf{c})$ of the query point $\mathbf{x}$ is calculated based on:
\begin{equation}
    (\sigma, \mathbf{c}) = (\sigma_{k^*}, \mathbf{c}_{k^*}), \text{where } k^* = \argmax\limits_{k}~ \sigma_k.
    \label{eq:aggregation}
\end{equation}

In contrast to \cite{peng2021animatable, liu2021neural} that represent the body with a single NeRF network, our part-based voxelized human representation can assign different densities of model parameters to different human parts with different complexity, thereby enabling us to efficiently distribute the representational power of the network. Experiments show that our proposed body representation significantly improves the rate of convergence.

\subsection{Training}
\label{sec:training}

The proposed representation can be learned from sparse multi-view videos by minimizing the difference between rendered and observed images. The volume rendering technique \cite{mildenhall2020nerf, kajiya1984ray} is used to synthesize the pixel color. Give a pixel at frame $t$, we emit a camera ray and sample points along the ray. Then, the sampled points are fed into the dynamic human representation to predict their colors and densities, which are finally accumulated into the pixel color. During each training iteration, we randomly sample an image patch from the input image and compute the Mean Squared Error (MSE) loss and perceptual loss \cite{johnson2016perceptual} to train the model parameters, which are defined as:
\begin{equation}
    L_{\text{rgb}} = \|\tilde{I}_P - I_P \|_2 + \|F_{\text{vgg}}(\tilde{I}_P) - F_{\text{vgg}}(I_P) \|_2,
\end{equation}
where $\tilde{I}_P$ is the rendered image patch, $I_P$ is the ground truth image patch, and $F_{\text{vgg}}$ extracts image features using the pretrained VGG network \cite{johnson2016perceptual}.
Ablation study demonstrates that perceptual loss is essential for rendering quality and fast training.

In addition to the image rendering loss, two regularization techniques are used to facilitate the learning of the neural representations.
First, we apply the regularizer in \cite{barron2021mip360} to concentrate densities on the human surface.
Second, the residual deformation field is regularized to be small and smooth.
More details of the loss terms are described in the supplementary material.

%% file: 04_implementation.tex
\subsection{Implementation details}
\label{sec:implementation}

We adopt the Adam optimizer \cite{kingma2014adam} with a learning rate of $5e^{-4}$.
We train our model on an RTX 3090 GPU, which takes around 5 minutes to produce photorealistic results.
Our method is implemented purely with the PyTorch framework \cite{paszke2019pytorch} to demonstrate the effectiveness of our representation
It also enables us to fairly compare with baseline methods \cite{peng2021neural, peng2021animatable, kwon2021neural} implemented in PyTorch.
The details of the network architectures and hyper-parameters are presented in the supplementary material.

%% file: 05_experiments.tex
\section{Experiments}

\subsection{Datasets}

\textbf{ZJU-MoCap \cite{peng2021neural} dataset} is a widely-used benchmark for human modeling from videos. It provides foreground human masks and SMPL parameters. Following \cite{weng2022humannerf}, we select 6 human subjects (377, 386, 387, 392, 393, 394) from this dataset to conduct our experiments. One camera is used for training, and the remaining cameras are used for evaluation. For each human subject, we select 1 frame every 5 frames and collect 100 frames for training. Please refer to the supplementary material for more detailed experiment settings of all characters.

\textbf{MonoCap dataset} contains four multi-view videos collected by \cite{peng2022animatable} from the DeepCap dataset \cite{habermann2020deepcap} and the DynaCap dataset \cite{habermann2021real}. It provides camera parameters and human masks. \cite{peng2022animatable} additionally estimate the SMPL parameters for each image. We adopt the setting of training and test camera views in \cite{peng2022animatable}. For each subject, 100 frames are selected for training, and we sample 1 frame every 5 frames. Detailed configurations of all sequences are described in the supplementary material.

\begin{figure*}[t]
    \centering
    \includegraphics[width=1\linewidth]{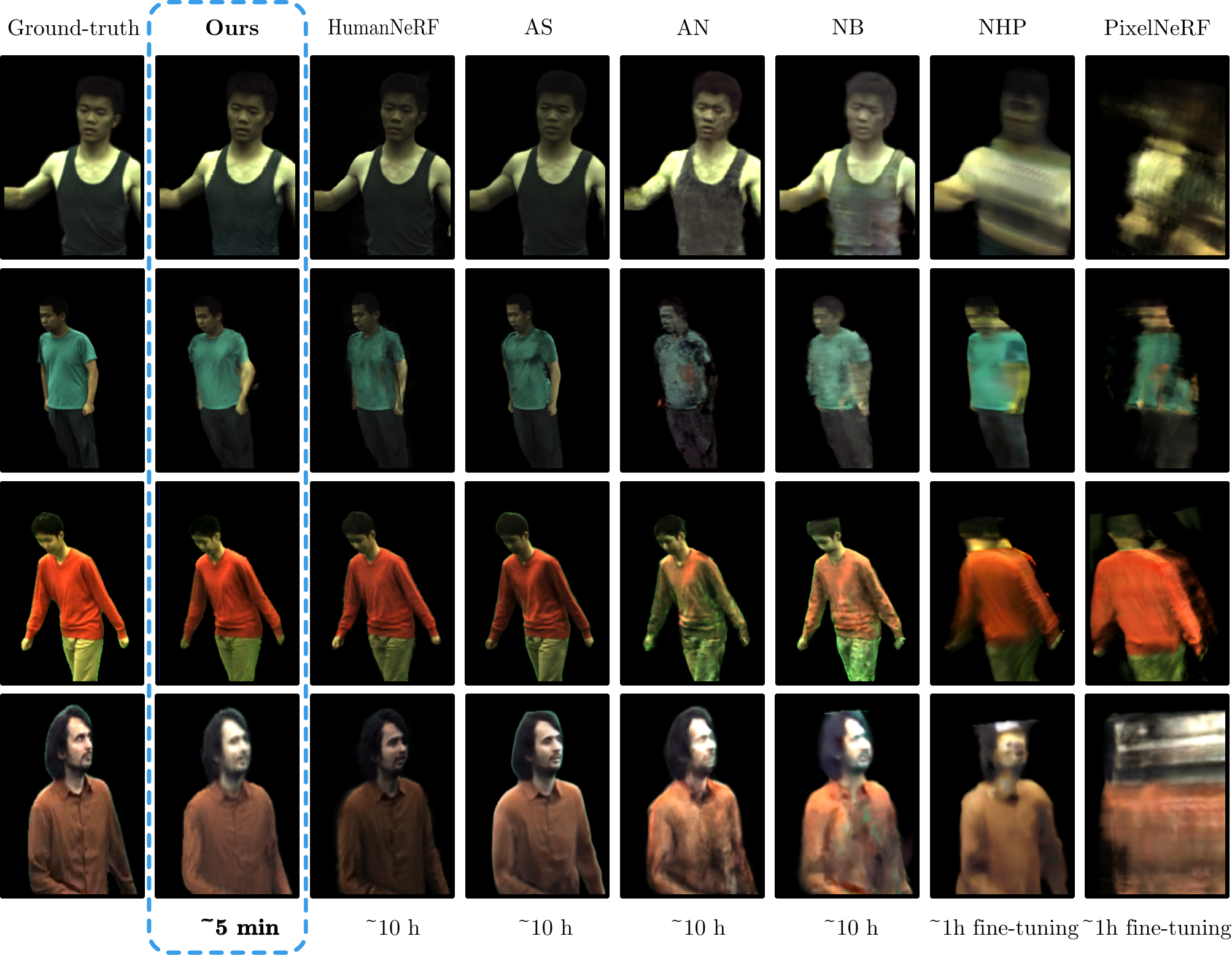}
    \caption{\textbf{Qualitative results of novel view synthesis on the ZJU-MoCap and MonoCap datasets.} Our method produces better rendering results while only requiring 1/100 of the training time. The bottom row lists the training time of each method.}
    \label{fig:comparison}
\end{figure*}

\subsection{Comparison with the state-of-the-art methods}

\paragraph{Baselines.} We compare our method with subject-specific optimization methods \cite{peng2021neural, peng2021animatable,weng2022humannerf, peng2022animatable}, generalizable methods \cite{yu2021pixelnerf, kwon2021neural}. All the baselines are implemented in pure PyTorch\cite{paszke2019pytorch} for a fair comparison. Here we list only the average metric values of all selected characters on a dataset due to the size limit. We provide more detailed qualitative and quantitative comparisons in the supplementary material.

(1) Subject-specific optimization methods. Neural Body (NB) \cite{peng2021neural} anchors a set of latent codes to the SMPL mesh and regresses the radiance field from the posed latent codes. Animatable NeRF (AN) \cite{peng2021animatable} deforms the canonical NeRF with the skeleton-driven framework and models non-rigid deformations by learning blend weight fields. \cite{peng2022animatable} extend \cite{peng2021animatable} with a signed distance field and pose-dependent deformation field to better model the residual deformation and geometric details of dynamic humans. \cite{weng2022humannerf} optimizes for a volumetric representation of the person in a canonical space along with the estimated human pose.

(2) Generalizable methods. PixelNeRF \cite{yu2021pixelnerf} trains a network to infer the radiance field from a single image. Neural Human Performer (NHP) \cite{kwon2021neural} anchors image features to vertices of the SMPL mesh and aggregates temporal features using a transformer, which are decoded into a human model. For each evaluated subject (e.g. one subject of MonoCap), we first pretrain the network on the other dataset (e.g. ZJU-MoCap) and then finetune it on the evaluated subject until it converges.

\input{figures/experiments/comparison}

\paragraph{Results on the ZJU-MoCap dataset.}

Table \ref{tab:comparison} compares our method with NB \cite{peng2021neural}, AN \cite{peng2021animatable}, PixelNeRF \cite{yu2021pixelnerf}, NHP \cite{kwon2021neural}, HN \cite{weng2022humannerf} and AS \cite{peng2022animatable} on novel view synthesis.
Our proposed representation can be optimized within around 5 minutes to produce photorealistic rendering results, while \cite{peng2021neural, peng2021animatable, weng2022humannerf, peng2022animatable} require around 10 hours to finish training and \cite{yu2021pixelnerf, kwon2021neural} require 10 hours of pretraining and 1 hour of fine-tuning.
\cite{weng2022humannerf,peng2022animatable} exhibit better results than \cite{peng2021neural, peng2021animatable}. However, they all require a lengthy optimization process and fail to produce reasonable renderings in only 5 minutes because their models have not converged yet.
Generalizable methods \cite{yu2021pixelnerf, kwon2021neural} failed to render humans with reasonable shapes under the monocular setting.
Our method achieves comparable results on all of the three evaluated metrics even when only trained in minutes, which shows the effectiveness of our novel human representation.
We present qualitative results of our method and baselines in Figure~\ref{fig:comparison}.

\paragraph{Results on the MonoCap dataset.}

Table~\ref{tab:comparison} summarizes the quantitative comparison between our method and other baselines on the MonoCap dataset. Our model again achieves competitive visual quality while only requiring 1/100 of the training time due to our efficient part-based voxelized human representation and effective motion parameterization scheme. Figure~\ref{fig:comparison} indicates that our method can produce better appearance details than \cite{peng2021neural, peng2021animatable, peng2022animatable}. Although \cite{peng2021neural, peng2021animatable} have shown impressive rendering results given 4-view videos, they do not perform well on monocular inputs. \cite{peng2021neural} implicitly aggregates the temporal information using structured latent codes, which may not work well on monocular videos with complex human motions. \cite{peng2021animatable} uses a learnable blend weight field to model human motion, which has a higher dimension and could be hard to converge well given single-view supervision. \cite{weng2022humannerf, peng2022animatable} demonstrate similar visual quality and show that representing the human motion with LBS model and residual deformation works particularly well, but their models require 100x more time to optimize.

\input{figures/experiments/ablation}

\begin{figure}[ht]
    \centering
    \includegraphics[width=0.7\linewidth]{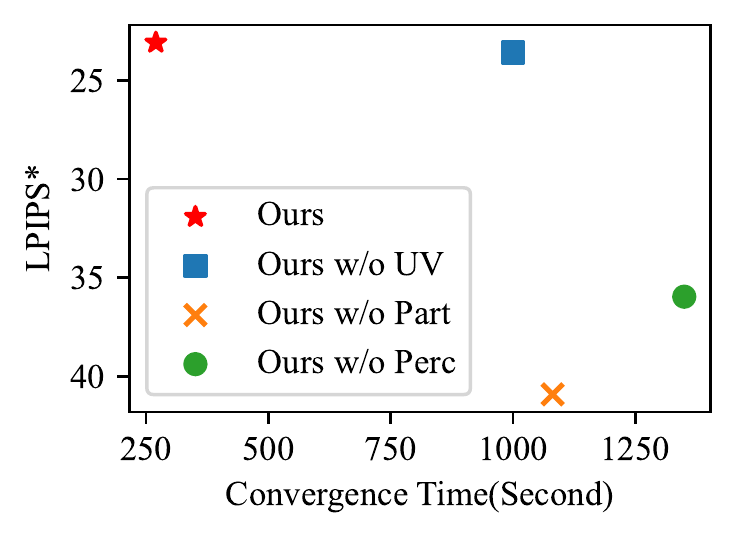}
    \caption{\textbf{Comparison of convergence LPIPS* and time needed for convergence of different variants of the proposed pipeline.} The description of each model can be found in Section \ref{sec:ablation}. The proposed components accelerate the training and enhance the rendering quality significantly. LPIPS*=LPIPS $\times 10^3$}
    \label{fig:ablation_plot}
    \vspace{-1em}
\end{figure}

\begin{figure*}[ht]
    \centering
    \includegraphics[width=0.8\linewidth]{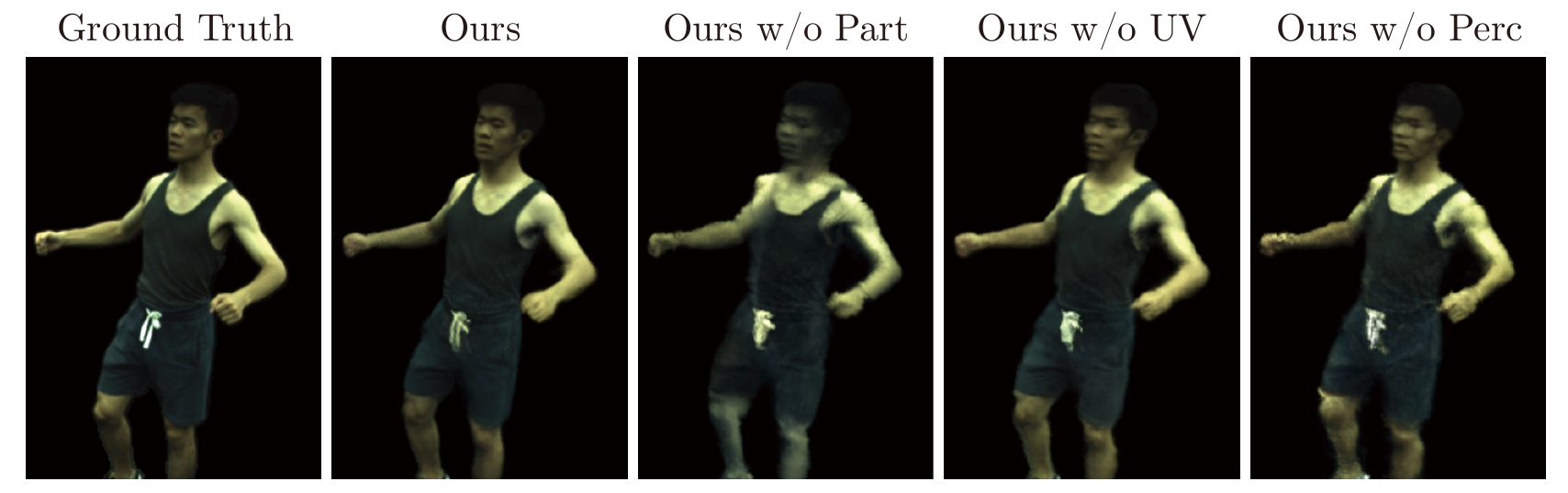}
    \caption{\textbf{Ablation studies on the 377 sequence of ZJU-MoCap dataset.} ``Ours w/o part" means that we use an MHE-augmented NeRF network to represent the whole body. ``Ours w/o UV" indicates that the residual deformation network $\text{MLP}_{\text{res}}$ takes the hash encoded $(\mathbf{x}, t)$ as input. ``Ours w/o Perc" represents that we do not use the perceptual loss during training.}
    \label{fig:ablation}
    \vspace{-1em}
\end{figure*}

\subsection{Ablation Studies}

\label{sec:ablation}

We perform ablation studies on the 377 sequence of the ZJU-MoCap dataset to analyze how the proposed components affect the performance and training speed of our method.

\subsubsection{Ablation Studies on Proposed Components.}

Table~\ref{tab:ablation}(a) lists the quantitative result of ablation studies on our proposed components. All models are trained for 5 minutes. "Ours w/o Part" represents the canonical human body with a single MHE-augmented NeRF \cite{muller2022instant} network, which drastically degrades the rate of convergence. To keep the comparison fair, this variant of our method has a similar number of parameters (302M) to ours (286M). However, this change leads to a significant decrease in PSNR of $1.98 \text{dB}$ because of its unwise design of considering all parts equally complex and wasting representational power. In "Ours w/o UV", the residual deformation network $\text{MLP}_{\text{res}}$ takes hash encoded $(\mathbf{x}, t)$ as input, which observed a PSNR degradation from $32.09 \text{dB}$ to $31.40 \text{dB}$ with the same training time because of severe hash collision and limited resolution. "Ours w/o Perc" does not adopt the perceptual loss during training, which in turn increases the LPIPS distance. This comparison illustrates the importance of perceptual loss for visual fidelity. \revise{Figure~\ref{fig:ablation_plot} and Figure~\ref{fig:ablation} provide a more intuitive qualitative comparison among variants of the proposed method.}

\paragraph{Analysis of the part-based voxelized human representation.}

MHE \cite{muller2022instant} defines a multiresolution hash table of trainable features to embed input coordinates to a high-dimension space. We find that simply increasing the size of the hash table does not always result in better performance with the same training time, because a bigger hash table leads to higher memory consumption and increases the time of each training iteration. The proposed part-based voxelized human representation allows us to adapt the hash table size according to the complexity of the human part, allowing us to efficiently represent the human body. Table~\ref{tab:ablation}(a) demonstrates the effectiveness of the part-based voxelized human representation. To further validate this representation, we design two variants that use the hash tables of size $2^{15}$ and $2^{20}$ in all human parts respectively. Table~\ref{tab:ablation}(c) summarizes the ablation studies, indicating that varying the model parameters in human parts improves the performance.

\paragraph{Analysis of the motion parameterization scheme.}

Table~\ref{tab:ablation}(a) shows that our model works better when the residual deformation network $\text{MLP}_{\text{res}}$ takes parameterized 3D surface-time $(u, v, t)$ coordinates as input, compared with taking 4D space-time $(\mathbf{x}, t)$ as input. Note that $(u, v, t)$ makes MHE much more memory efficient than $(\mathbf{x}, t)$. To further validate the effectiveness of our motion parameterization, we additionally design three variants of the $\text{MLP}_{\text{res}}$ input. (1) PE: positional encoded $(\mathbf{x}, t)$. (2) XYZ-Code: hash encoded $\mathbf{x}$ and a per-frame learnable latent code \cite{peng2022animatable}. (3) XYZ-Pose: hash encoded $\mathbf{x}$ and the pose parameter \cite{liu2021neural, weng2022humannerf}. When taking the positional encoded $(\mathbf{x}, t)$ as input, we use a larger network for $\text{MLP}_{\text{res}}$. The results in Table~\ref{tab:ablation}(b) indicate that hash encoded $(u, v, t)$ achieves the best performance.

\revise{\paragraph{Analysis of Robustness.} To evaluate the robustness of the proposed system, we measure the time needed to achieve an evaluation PSNR of 30 for five times on ``377'' sequence. This results in a training time with a mean value of $76.00s$ and a standard derivation of $13.56s$, showing the stability of the proposed method.}

%% file: figures/experiments/comparison.tex
\begin{table*}

    \centering

    \resizebox{0.8\textwidth}{!}{%
        \begin{tabular}{c|c|ccc|ccc}

            \hline
            \multicolumn{1}{l|}{}             & \multicolumn{1}{l|}{}                                & \multicolumn{3}{c|}{ZJU-MoCap}      & \multicolumn{3}{c}{MonoCap}                                                                                                                                                                                \\
            \textbf{}                         & \multicolumn{1}{l|}{\multirow{-2}{*}{Training Time}} & \multicolumn{1}{c}{PSNR $\uparrow$} & \multicolumn{1}{c}{SSIM $\uparrow$} & \multicolumn{1}{c|}{LPIPS$^*$ $\downarrow$} & \multicolumn{1}{c}{PSNR $\uparrow$} & \multicolumn{1}{c}{SSIM $\uparrow$} & \multicolumn{1}{c}{LPIPS$^*$ $\downarrow$} \\

            \hline
            \rowtopspace
            Ours                              & \textbf{\textasciitilde 5 min}                                     & \textbf{31.01}                      & \underline{0.971}                   & 38.45                                       & \underline{32.61}                   & \textbf{0.988}                      & 16.68                                      \\
            HumanNeRF\cite{weng2022humannerf} & ~10 h                                                & \underline{30.66}                   & 0.969                               & \textbf{33.38}                              & \textbf{32.68}                      & \underline{0.987}                   & \underline{15.52}                          \\
            AS\cite{peng2022animatable}       & \textasciitilde 10 h                                                & 30.38                               & \textbf{0.975}                      & \underline{37.23}                           & 32.48                               & \textbf{0.988}                      & \textbf{13.18}                             \\
            AN\cite{peng2021animatable}       & \textasciitilde 10 h                                                & 29.77                               & 0.965                               & 46.89                                       & 31.07                               & 0.985                               & 19.47                                      \\
            NB\cite{peng2021neural}           & \textasciitilde 10 h                                                & 29.03                               & 0.964                               & 42.47                                       & 32.36                               & 0.986                               & 16.70                                      \\
            NHP\cite{kwon2021neural}          & \underline{\textasciitilde 1 h fine-tuning}                          & 28.25                               & 0.955                               & 64.77                                       & 30.51                               & 0.980                               & 27.14                                      \\
            \rowbotspace
            PixelNeRF\cite{yu2021pixelnerf}   & \underline{\textasciitilde 1 h fine-tuning}                          & 24.71                               & 0.892                               & 121.86                                      & 26.43                               & 0.960                               & 43.98                                      \\
            \hline
        \end{tabular}%
    }
    \vspace{1em}
    \caption{\textbf{Quantitative comparison} of our method and baseline methods on the ZJU-MoCap and MonoCap datasets. We use \textbf{bold text} for the best and \underline{underlined text} for the second best metric value across methods. Our method achieves the fastest training speed and shows competitive rendering results. Note that the NHP \cite{kwon2021neural} and PixelNeRF \cite{yu2021pixelnerf} are additionally pretrained for 10 hours. LPIPS$^*$ = LPIPS $\times 10^3$.}
    \label{tab:comparison}
\end{table*}

%% file: figures/experiments/ablation.tex
\begin{table*}
    \centering

    \resizebox{0.32\textwidth}{!}{
        \subfloat[Ablation studies on proposed components.]{
            \begin{tabular}{clll}
                \hline
                              & PSNR           & SSIM           & LPIPS$^*$      \\ \shline
                \rowtopspace
                Ours          & \textbf{32.09} & \textbf{0.982} & \textbf{23.47} \\
                Ours w/o Part & 30.11          & 0.974          & 45.84          \\
                Ours w/o UV   & 31.40          & 0.979          & 30.99          \\
                \rowbotspace
                Ours w/o Perc & 30.55          & 0.976          & 44.33          \\ \hline
            \end{tabular}
            \label{tab:ablation_components}
        }
    }
    \resizebox{0.30\textwidth}{!}{
    \subfloat[Ablation studies on variants of $\text{MLP}_{\text{res}}$ input.]{
    \begin{tabular}{clll}
        \hline
                 & PSNR           & SSIM           & LPIPS$^*$      \\ \shline
        \rowtopspace
        Ours     & \textbf{32.09} & \textbf{0.982} & \textbf{23.47} \\
        PE       & 31.94          & 0.981          & 26.76          \\
        XYZ-Code & 31.32          & 0.979          & 31.19          \\
        \rowbotspace
        XYZ-Pose & 31.51          & 0.979          & 34.42          \\ \hline
    \end{tabular}
    \label{tab:ablation_motion_field}
    }
    }
    \resizebox{0.3175\textwidth}{!}{
        \subfloat[Ablation studies on the part parameters.]{
            \begin{tabular}{cccc}
                \hline
                                    & PSNR           & SSIM           & LPIPS$^*$      \\ \shline
                \rowtopspace
                Ours                & \textbf{32.09} & \textbf{0.982} & \textbf{23.47} \\
                Table size $2^{15}$ & 30.69          & 0.976          & 35.67          \\
                Table size $2^{20}$ & 31.18          & 0.978          & 33.58          \\
                \rowbotspace
                                    &                &                &                \\
                \hline
            \end{tabular}
            \label{tab:ablation_part}
        }
    }

    \vspace{1em}
    \caption{\textbf{Quantitative results of ablation studies} on the 377 sequence of ZJU-MoCap dataset. The description of each model can be found in Section~\ref{sec:ablation}. All models are trained for 5 minutes.}
    \label{tab:ablation}
\end{table*}

%% file: 06_limitations.tex
\section{Limitations}

Although our method can quickly reconstruct high-quality human models from videos, there are still some challenges. %
First, it is desirable to infer human models from sparse multi-view images.
Improving generalizable methods like \cite{yu2021pixelnerf, kwon2021neural} with our proposed components may be a direction to solve this problem.
Second, our method currently relies on accurate SMPL parameters, which could be difficult to obtain under in-the-wild settings. It is interesting to utilize the techniques in \cite{su2021nerf, weng2022humannerf} to optimize the human pose parameters along with the training of human avatars.
Third, we can only reconstruct foreground dynamic humans, while dynamic scenes typically include foreground and background entities. It might be plausible to combine our method with ST-NeRF \cite{zhang2021stnerf} to quickly reconstruct dynamic scenes containing foreground and background objects.

%% file: 10_conclusion.tex
\label{sec:conclusion}

\section{Conclusion}

We introduced a novel dynamic human representation that can be quickly optimized from videos and used for generating free-viewpoint videos of the human performer.
This representation consists of a part-based voxelized human model in the canonical space and a motion parameterization scheme that transforms points from the world space to the canonical space.
The part-based voxelized human model decomposes the human body into multiple parts and represents each part with an MHE-augmented NeRF network, which efficiently distributes network representational power and significantly improves the training speed.
When predicting the motion of a query point, the motion field reparameterizes the point coordinate as 2D surface-level UV coordinate, which effectively reduces the dimensionality of motion the network is required to model, resulting in a boost in convergence rate.
Experiments demonstrate that our proposed representation can be optimized at 1/100 of the time of previous methods while still maintaining competitive rendering quality.
We show that given a 100-frame monocular video of $512 \times 512$ resolution, our method can produce photorealistic free-viewpoint videos in minutes on an RTX 3090 GPU.